\documentclass{ecai} 

\usepackage{latexsym}
\usepackage{amssymb}
\usepackage{amsmath}
\usepackage{amsthm}
\usepackage{booktabs}
\usepackage{enumitem}
\usepackage{graphicx}
\usepackage{color}
\usepackage{multirow}

\newcommand{\BibTeX}{B\kern-.05em{\sc i\kern-.025em b}\kern-.08em\TeX}

\begin{document}


\begin{frontmatter}


\paperid{123} 


\title{AHMsys: An Automated HVAC Modeling System for BIM Project}


\author[A,B]{\fnms{Long Hoang}~\snm{Dang}\thanks{Corresponding Author. Email: long.dang@akila3d.com}}
\author[A]{\fnms{Duy-Hung}~\snm{Nguyen}}
\author[A]{\fnms{Thai Quang}~\snm{Le}}
\author[A]{\fnms{Thinh Truong}~\snm{Nguyen}}
\author[A]{\fnms{Clark}~\snm{Mei}}
\author[A]{\fnms{Vu}~\snm{Hoang}}

\address[A]{Akila}
\address[B]{Posts and Telecommunications Institute of Technology}


\begin{abstract}
This paper presents a novel system, named AHMsys, designed to automate
the process of generating 3D Heating, Ventilation, and Air Conditioning
(HVAC) models from 2D Computer-Aided Design (CAD) drawings, a key
component of Building Information Modeling (BIM). By automatically
preprocessing and extracting essential HVAC object information then
creating detailed 3D models, our proposed AHMsys significantly reduced the 20\% work schedule of the BIM process in Akila. This advancement highlights the essential impact of integrating
AI technologies in managing the lifecycle of a building's digital
representation.
\end{abstract}

\end{frontmatter}


\section{Introduction}

Building Information Modeling (BIM) is a comprehensive technology
that involves the creation and maintenance of a digital representation
(model) of a physical structure throughout its entire lifecycle, becoming
crucial in the field of construction. This model is more than just
a collection of geometric shapes; it includes detailed information
about the components, properties, and spatial relationships
of the objects in the building. Among its various aspects, Heating,
Ventilation, and Air Conditioning (HVAC) modeling - the process that
generate a 3D model from 2D Computer-Aided Design (CAD) drawings,
stands out as a critical and often the most labor-intensive phase
in a BIM project \cite{buildings14030788}. With the rapid development of Artificial Intelligence
(AI), there is an urgent and real need in architectural and construction
companies to develop a system that can fully automate the HVAC-modeling
stage, which can improve project efficiency, reduce human labors and
production costs, and enhancing market competitiveness. 

\begin{figure}[t]
\includegraphics[width=0.9\columnwidth]{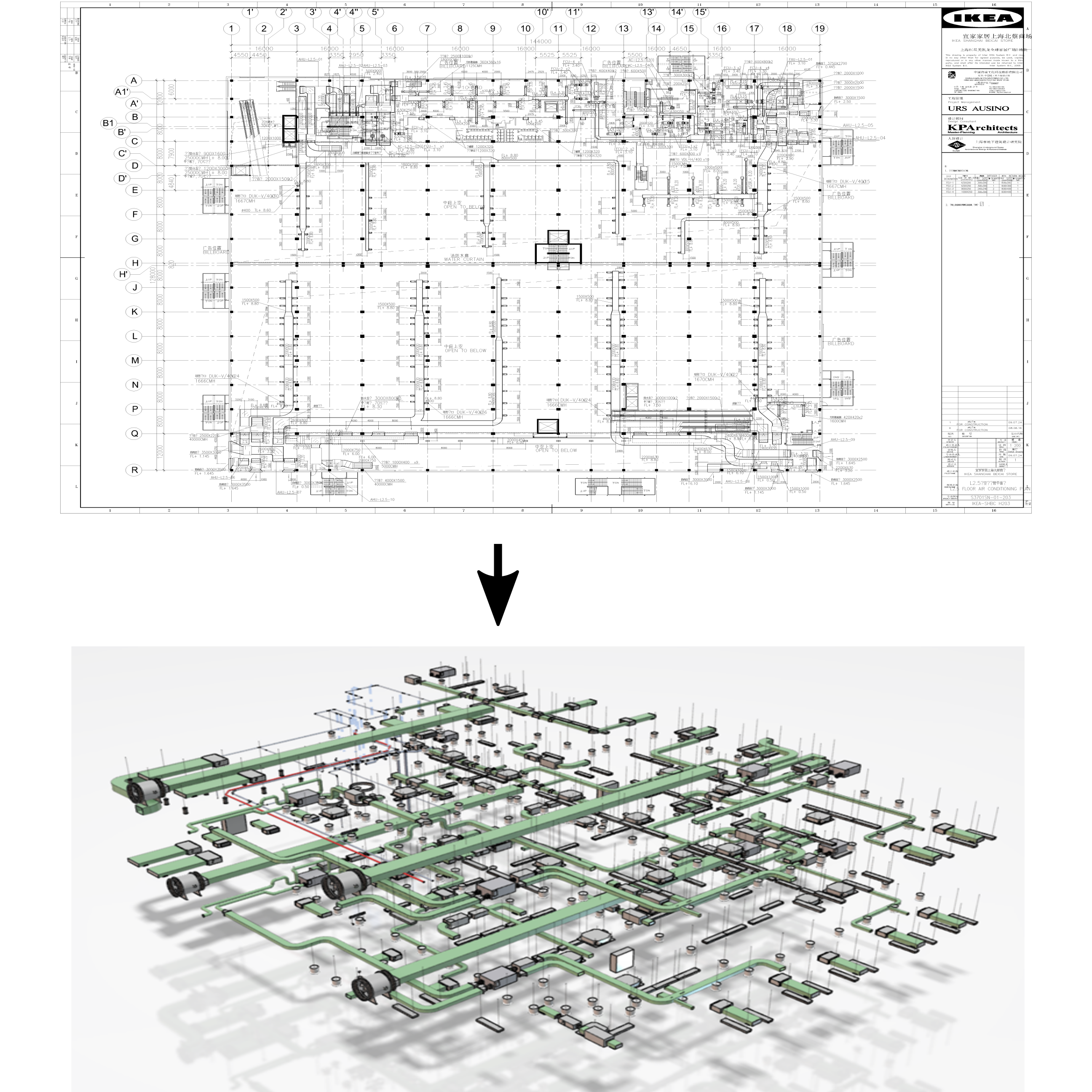}

\caption{The 3D HVAC model generated by our proposed AHMsys based on the 2D
CAD drawing. \label{fig:Teaser}}
\end{figure}

Previous studies have frequently utilized deep learning models, including
CNNs \cite{Rezvanifar2020SymbolSO,Fu2011FromED,Sarkar2022AutomaticDA,Rusiol2010SymbolSI},
GAN \cite{Elyan2020DeepLF}, and Graph Neural Networks (GNNs) \cite{8893047}
to automate the extraction of information from CAD drawings. These
approaches typically necessitate large amounts of annotated data for
training and significant computational resources \cite{Elyan2018SymbolsCI}.
Unlike these approaches, our work looks into a lightweight system
designed to be more accessible for the majority of architecture and
construction companies, which often lack the extensive labeled data
and machines available to larger entities. Furthermore, to the best of our knowledge, our project is the first to offer an automated solution that does not require extensive labeled data or significant computational resources, capable of directly converting 2D CAD drawings into 3D HVAC models. Figure \ref{fig:Teaser} illustrates a 3D model of an HVAC system generated
by our proposed method.

\begin{figure*}[t]
\includegraphics[width=1\textwidth]{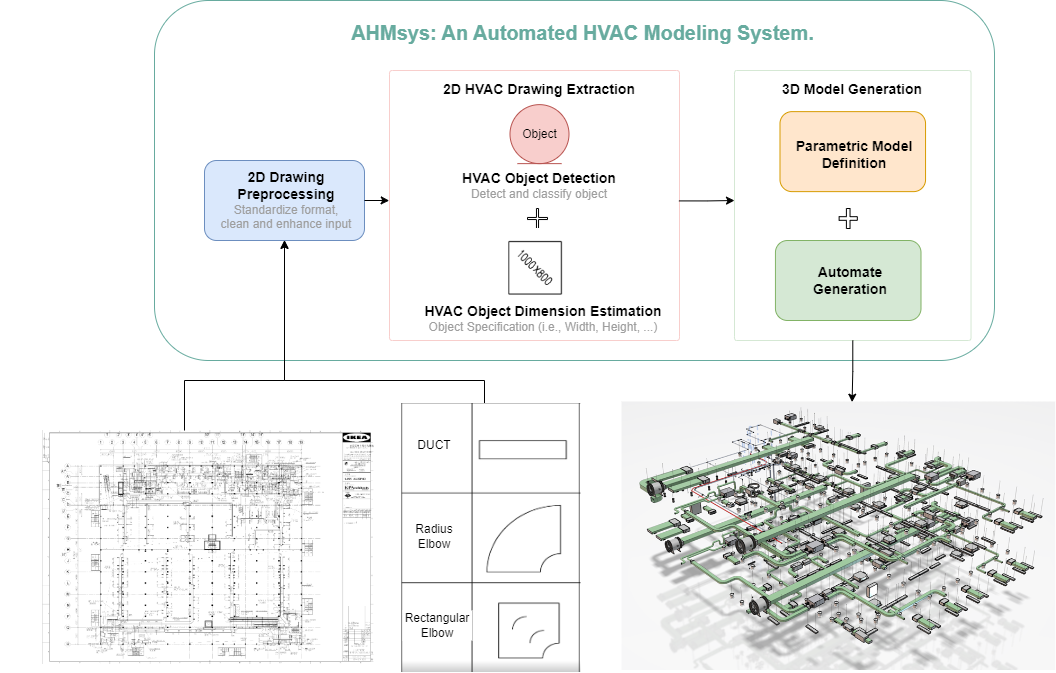}

\caption{Our proposed AHMsys contains three key steps: 1) 2D Drawing Preprocessing,
2) 2D HVAC Drawing Extraction, and 3) 3D Model Generation. Given a
2D CAD drawing and a list of potential object types within the drawing
as inputs, this framework will create integrated 3D HVAC model as
output. \label{fig:Overview}}
\end{figure*}

In this paper, we introduce AHMsys: Automated HVAC Modeling System,
designed to fully automate the HVAC modeling process within BIM projects.
Our framework is structured around three principal phases: 1) 2D Drawing
Preprocessing, 2) 2D HVAC Drawing extraction, and 3) 3D Model Generation.
We have effectively developed and incorporated AI technologies that
are well-suited for operations on a tight budget, making it suitable
for small to medium-sized companies to adopt easily. AHMsys has been
successfully implemented in Akila, a global firm offering ESG solutions
for buildings and factories, where it has substantially reduced the work schedule 20\%
of BIM projects that were previously dependent on manual labor.


\section{System Framework}

We present an overview of our proposed AHMsys in Figure \ref{fig:Overview}.
This framework takes a 2D CAD drawing and a list of potential object
types within the drawing as inputs, and produces an integrated 3D
model as output. It is important to note that the primary objective
of this work is to generate 3D models specifically for HVAC systems.
As such, we preprocess the input by removing non-HVAC objects from
the drawing. The workflow comprises several key steps: Firstly, drawings
are exported from CAD software and undergo preprocessing to prepare
them for subsequent analysis. Subsequently, the 2D HVAC Drawing Extraction
module identifies and extracts essential information about each object,
including its type, specifications, and spatial coordinates. Finally,
the 3D Model Generation engine utilizes this extracted information
to construct a comprehensive 3D model. Each component of this process
will be discussed in further detail in the sections that follow.

\subsection{2D Drawing Preprocessing}

The objective of the data preprocessing step is to refine the drawings
to a high standard of quality, thereby optimizing them for next analytical
processes. Initially, 2D CAD drawings are cleansed of non-HVAC objects
and adhere to the following standards. The images are exported as
transparent PNGs to maintain visual clarity. These images must meet
an 8K UHD resolution (7680x4320 pixels) to ensure the highest level
of detail. A monochrome color scheme (black and white) is applied
to simplify the identification process. Within these images, HVAC
objects are distinguished by a line thickness of 0.6mm, while other
elements are represented with a 0.15mm thickness. The dimensions of
each HVAC object are annotated in a textbox positioned directly next
to the object. A reference node, designed as an equilateral triangle
with dimensions of 500x500x500mm, is placed at a known coordinate
or the origin of the CAD drawing. This reference node aids in the
accurate calculation of the real image scale. Additionally, a selected
list of the most common HVAC object types is prepared. This list serves
to enhance the accuracy of the proposed system by ensuring that the
most relevant objects are identified and analyzed.

\subsection{2D HVAC Drawing Extraction}

\subsubsection{HVAC Object Detection}

As mentioned earlier, the scarcity of 2D HVAC drawings available for
training deep learning models in existing work presents a notable
limitation. Additionally, many construction and architecture firms
lack the requisite computational resources to deploy deep learning
models effectively. To address these challenges, we have employed
Suzuki\textquoteright s Contour tracing algorithm \cite{Suzuki1985TopologicalSA}
in OpenCV library \cite{opencv_library} to identify HVAC objects
within the drawings and calculate their spatial coordinates.

Our analysis revealed that Ducts are the most critical objects within
these drawings (see Table 1). Therefore, we focus on recognizing Ducts based on
a list of common HVAC object types, proceeding to calculate their
nine spatial coordinates. These coordinates include the four corner
points, four midpoints, and the center point of the Ducts. The determination
of the Ducts' direction is achieved by examining the mid-coordinates
that intersect with other objects, facilitating the identification
of continuous Ducts and their neighboring points along the presumed
direction. In the case where a Duct is not continuous, and proximity
to a reference point is established, the detection of other objects
is guided by the following rule: a single nearest point plus space
indicates an Elbow; two nearest points plus space suggest a Tee; and
three nearest points plus space denote a Cross.

\subsubsection{HVAC Object Dimension Estimation}

To generate an accurate 3D model for a HVAC system, it is crucial
not only to identify the type and location of each object but also
to accurately estimate its dimensions. Initially, we employ the Suzuki\textquoteright s
Contour algorithm to efficiently detect the bounding boxes of text
areas. Subsequently, the cropping function in OpenCV is utilized to
segment the image according to these boundary boxes. Then, PaddleOCR
\cite{Du2020PPOCRAP}, an open-source library, is leveraged for text
recognition on the segmented images. In the final step, we integrate
detailed object dimensions with their respective types and locations
to ensure the generation of highly accurate 3D models.

\subsection{3D Model Generation}

The purpose of our 3D generation module is to create a detailed and
accurate 3D models based object information extracted in the 2D HVAC
drawing extraction stage. The process leverages a two-step approach:
(1) Parametric Model Definition: we utilize the extracted data to
define key parameters of the 3D model, such as dimensions, shapes,
orientations, and material properties. These parameters are established
within a pre-defined parametric model, essentially serving as the
building blocks for constructing the 3D geometry. (2) Automated Generation:
to achieve full automation, we developed a tool named \textquotedbl Automation
Scripts\textquotedbl . Once the parameters are set, this script automatically
generates the 3D model.


\section{Results}

We evaluate the effectiveness of our proposed AHMsys using a dataset
of 2D CAD drawings collected from a BIM project at Akila in 2022.
This dataset contains 63,629 HVAC objects from 110 CAD drawings. It
has been determined that the most crucial object types in CAD drawings
are Duct, Elbow and Tee, which collectively appear in nearly 86\%
of cases. The detailed statistics of our dataset are presented in
Table \ref{tab:statistics}.

\begin{table}
\begin{tabular}{c|c|c|c|c|c}
\hline 
\# CAD drawings & \# Objects & \# Duct & \# Elbow & \# Tee & \# Other\tabularnewline
\hline 
110 & 63,629 & 33,723 & 8,908 & 12,089 & 8,909\tabularnewline
\hline 
\end{tabular}

\caption{Statistics of our experimental dataset. \# stands for \textquotedbl Number
of\textquotedbl{} \label{tab:statistics} }
\end{table}

Table \ref{tab:Performance-of-AHMsys} presents the performance of
AHMsys on our 2D CAD drawing dataset. As shown, the results demonstrate
that the 3D models generated by AHMsys precisely represent the type,
location, and dimensions of HVAC objects, achieving accuracy scores
of 98.9\%, 98.1\%, and 98.4\% respectively. Notably, AHMsys achieves
an accuracy of over 98\% when modeling Ducts, which are among the
most popular and critical objects in the CAD drawings. These results
and analysis are clear evidence of the effectiveness of AHMsys in
performing the HVAC modeling process within BIM projects.

\begin{table}
\begin{tabular}{l|l|l|l|l|c}
\hline 
\multirow{2}{*}{} & \multicolumn{5}{c}{Accuracy$\uparrow$}\tabularnewline
\cline{2-6} \cline{3-6} \cline{4-6} \cline{5-6} \cline{6-6} 
 & Duct & Elbow & Tee & Other & All\tabularnewline
\hline 
Type of objects & 99.6\% & 96.7\% & 99.2\% & 98\% & \multicolumn{1}{c}{98.9\%}\tabularnewline
Location of objects & 98.8\% & 95.1\% & 99\% & 97.3\% & 98.1\%\tabularnewline
Dimensions of objects & 99.1\% & 95.8\% & 99.2\% & 97.2\% & 98.4\%\tabularnewline
\hline 
\end{tabular}\caption{Performance of AHMsys on Akila's 2D CAD drawing dataset \label{tab:Performance-of-AHMsys}}
\end{table}

\begin{figure}
\includegraphics[width=0.9\columnwidth]{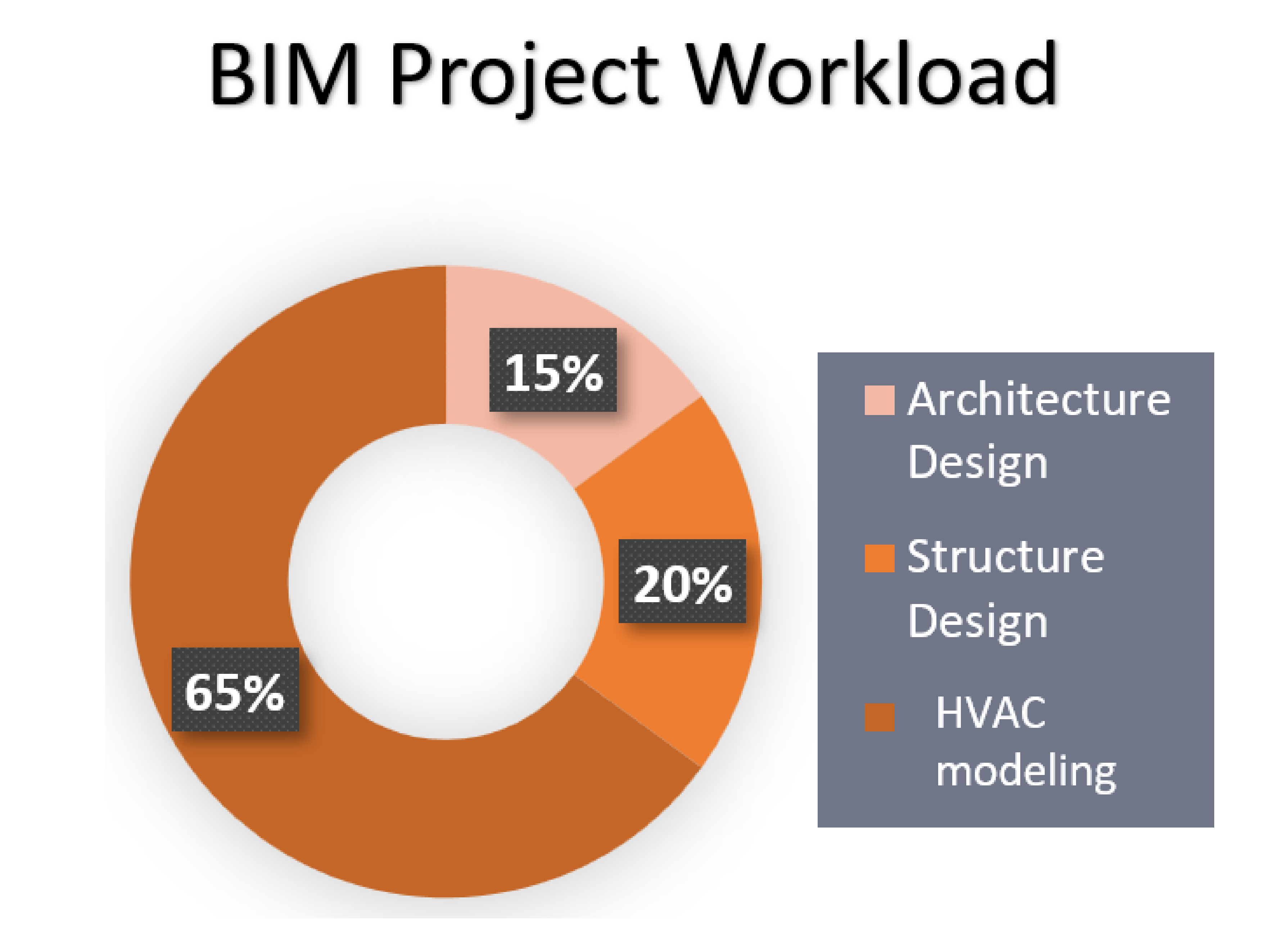}

\caption{The workload for the BIM project at Akila Company in 2022 was mainly
focused on HVAC modeling, which accounted for 65\% of the total. The
remainder was distributed among other disciplines, including architecture
and structure design. \label{fig:The-workload}}

\end{figure}

Additionally, we present data on the workload (see Figure \ref{fig:The-workload})
and the project management report for the BIM project at Akila in
2022, which was deployed across three different cities: Shanghai,
Chengdu, and Suzhou (see Table \ref{tab:The-project-management}).
It is clear that implementing our proposed AHMsys significantly reduced
the work schedule by 20\% compared to the estimated working hours
when tasks were conducted entirely manually by BIM engineers. These
results highlight the effectiveness of AHMsys in automating the BIM
process and enhancing the company's profitability.

\begin{table}
{\scriptsize{}}%
\begin{tabular}{l|c|c|c}
\hline 
{\scriptsize{}Cities} & {\scriptsize{}\# BIM Engineers} & {\scriptsize{}Est. working hours } & {\scriptsize{}Actual working hours}\tabularnewline
\hline 
{\scriptsize{}Shanghai} & {\scriptsize{}3} & {\scriptsize{}773} & {\scriptsize{}626}\tabularnewline
{\scriptsize{}Chengdu} & {\scriptsize{}4} & {\scriptsize{}707} & {\scriptsize{}552}\tabularnewline
{\scriptsize{}Suzhou} & {\scriptsize{}3} & {\scriptsize{}309} & {\scriptsize{}253}\tabularnewline
\hline 
{\scriptsize{}All} & {\scriptsize{}10} & {\scriptsize{}1789} & {\scriptsize{}1431}\tabularnewline
\hline 
\end{tabular}{\scriptsize\par}

\caption{The project management report for the BIM project at Akila \label{tab:The-project-management}}
\end{table}

\section{Conclusion}

In this paper, we develop a novel system named AHMsys that automatically
generates 3D HVAC model from 2D drawing for BIM project. Our proposed
AHMsys includes three main aspects, aiming at 2D Drawing Preprocessing,
2D HVAC Drawing Extraction, and 3D Model Generation, respectively.
Notably, our proposed framework represents a completely novel and
innovative approach within the construction domain, optimize processes
traditionally performed by BIM engineers. Currently deploying at our
company, Akila, it has significantly reduced the workload and production
costs associated with BIM projects.

In the future, we aim to improve our automated HVAC modeling system
by incorporating additional development frameworks. Moreover, we intend
to expand the capabilities of our tool to manage more complex elements
within buildings, such as piping and electrical components.



\bibliography{mybibfile.bib}

\end{document}